# Dual-attention Focused Module for Weakly Supervised Object Localization

**Yukun Zhou, Zailiang Chen, Hailan Shen\*, Qing Liu, Rongchang Zhao, and Yixiong Liang**

School of Computer Science and Engineering, Central South University, Hunan, Changsha 410083, China
hn_shl@126.com

**Abstract**

The research on recognizing the most discriminative regions provides referential information for weakly supervised object localization with only image-level annotations. However, the most discriminative regions usually conceal the other parts of the object, thereby impeding entire object recognition and localization. To tackle this problem, the Dual-attention Focused Module (DFM) is proposed to enhance object localization performance. Specifically, we present a dual attention module for information fusion, consisting of a position branch and a channel one. In each branch, the input feature map is deduced into an enhancement map and a mask map, thereby highlighting the most discriminative parts or hiding them. For the position mask map, we introduce a focused matrix to enhance it, which utilizes the principle that the pixels of an object are continuous. Between these two branches, the enhancement map is integrated with the mask map, aiming at partially compensating the lost information and diversifies the features. With the dual-attention module and focused matrix, the entire object region could be precisely recognized with implicit information. We demonstrate outperforming results of DFM in experiments. In particular, DFM achieves state-of-the-art performance in localization accuracy in ILSVRC 2016 and CUB-200-2011.

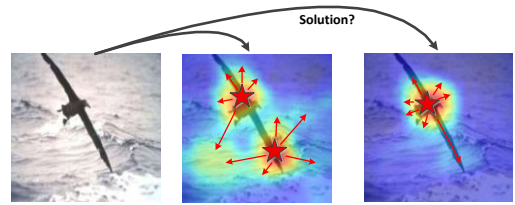

Fig 1: Attention transfers to the background with ADL method. The red arrows indicate the transfer direction and the stars mean the most discriminative regions. Our work is to put forward a solution for accurate object attention computation and object localization.

## Introduction

In recent years, object detection has achieved satisfactory performance in recent years (Ren et al. 2017; Zhou et al. 2019; Lin et al. 2017). However, it requires abundant location annotations work which is time-consuming and tedious. Weakly Supervised Objective Localization (WSOL) algorithms are devised to alleviate the obstacle. WSOL algorithms are able to detect the object regions with only the image-level labels instead of the bounding box information.

In some researches, scholars discover the discriminative features in classification network are significant reference information which helps locate the objects. Unfortunately, the CAM merely focuses on the most discriminative regions (Zhou et al. 2016).

To solve this problem, Hide-and-Seek method is proposed to hide patches in a training image randomly, forcing the network to seek other relevant parts (Singh and Lee, 2017). Wei et al. drive the classification network to sequentially discover new and complement object regions by erasing the current mined regions in an adversarial manner (Wei et al. 2017). Zhang et al. utilize two adversary classifiers for complementary object regions localization. One classifier is employed to guide the erasing operation and the other one discovers the rest parts of object regions (Zhang et al. 2018). A guided attention inference network is proposed to treat the most discriminative parts as masks and search the other regions in the following network (Li et al. 2018). To simplify the network structure and keep multi-time object regions extraction, Attention-based Dropout Layer (ADL) is presented to improve the accuracy of WSOL (Choe and Shim 2019). It completes the most discriminative regions erasing in the dropout layer which enhances computing efficiency. The importance map is utilized to increase the classification accuracy.

Although the most discriminative regions erasing is an effective idea for entire object detection, there are two drawbacks for the present researches. 1) The erasing methods abandon all the information on the most discriminative regions, thereby sometimes resulting in attention misdirection, which leads to biased localization. For example, as shown in Fig. 1, the attention regions spread to the background with ADL method. In this case, the bounding box is too large to precisely locate the object. 2) The classification accuracy has declined to some degree mainly in two reasons. Firstly, the classification network structures have been altered, so the classification performance is not as good as before. Second, this issue also relates to the attention misdirection problem, which leads to a wrong classification as the focused attention has been changed to other objects. In order to tackle these two challenges, we propose a Dual-attention Focused Module (DFM) to enhance object classification and localization performance.

DFM inherits the two advantages of the ADL algorithm. It needs no extra trainable parameters compared to the baselines, so it is weight light. Additionally, the module is insertable as it could be applied in kinds of CNN. On the other side, the differences between proposed DFM and ADL are mainly in two marked parts. Firstly, DMF considers about the attention regions in more comprehensive aspects. In DFM, we present a dual-attention module, consisting of a *position branch* and a *channel branch*. For *channel branch*, a Channel Enhancement Map (CEM) and a Channel Mask Map (CMM) are computed. The CEM is utilized to distribute large weights on the discriminative channels and little weights on the valueless ones, which could improve the classification accuracy. The CMM conceals the most informative channels so that the subsidiary important ones could be highlighted in the following layers computation. For *position branch*, there is the similar process, but the information dimension is on the position, so a Position Enhancement Map (PEM) and a Position Mask Map (PMM) are acquired.

What the most significant is that we complete the information fusion and complementary. Specifically, the CEM is summed with PMM, while CMM is combined with PEM. The CEM could supplement the abandoned information of PMM, so the attention extraction in the following layers will not be scattershot as CEM provides the weighted implicit information to the next layer. Hence, the object localization accuracy could be increased. The added implicit information behaves better than directly maintaining partial the most discriminative parts, because it includes the channel significance information which furthest improves the classification accuracy. Similarly, the PEM offers the CMM implicit position information during the subsidiary important channel extraction. In addition, we design a neighbor focused matrix to further avoid attention deviation. It utilizes the basic principle that the pixels of an object are continuous. We strengthen the mask surrounding pixels value to certain multiples to prevent attention from being diverted to non-object.

In general, there are three contributions of DFM to WSOL.

1) We reveal two drawbacks of the present methods based on the most discriminative regions erasing, influencing the WSOL performance. To address these problems, the DFM is proposed to enhance object classification and localization accuracy.
2) Two branches are employed to include comprehensive information. The information fusion between the *position branch* and the *channel branch* is presented. It provides the implicit information to make up the disadvantage of abandoning the most discriminative regions in mask maps, and improve the classification accuracy, thereby compensating the two drawbacks.
3) The neighbor focused matrix is proposed to keep extracted attention continuously gathering on the object, auxiliary improving the WSOL performance. This method has achieved state-of-the-art effects in CUB-200-2011 and ILSVRC 2016.

## Related Works

### Attention Mechanism

Attention mechanism is an effective data processing method learnt from human perception process (Mnih et al. 2014). It decides the distribution of available processing resources, which places more weights on the most informative regions (Mnih et al. 2014; Bahdanau, Cho, and Bengio 2014). With this feature, it brings many benefits to various fields, such as image captioning (Xu et al. 2015), image localization and understanding (Cao et al. 2015; Jaderberg et al. 2015; Zhou et al. 2016; Li et al. 2018; Wei et al. 2018), image inpainting (Yu et al. 2018; Liu et al. 2018), scene segmentation (Fu et al. 2019). Wang et al. propose a trunk-and-mask attention mechanism using an hourglass module (Wang et al. 2017). Squeeze-and-excitation block is proposed in (Hu, Shen, and Sun 2018). It is specialized to model channel-wise relationships in a computationally efficient manner and designed to enhance the representational power. Woo et al. present Convolutional Block Attention Module (CBAM) to utilize the channel attention and spatial attention at the same time, increasing the accuracy of the object classification (Woo et al. 2018). An attention branch is proposed for classification performance improvement (Fukui et al. 2019), which extends a response-based visual explanation model by introducing a branch structure with an attention mechanism. A Dual Attention Network (DANet) is proposed to build two global dependencies for scene segmentation (Fu et al. 2019). Compared with CBAM and DANet, the proposed FDM generates the mask maps to improve object localization. Meanwhile, the *position branch* and the *channel branch* are fully self-attention without any overhead parameters, and the classification and localization accuracy are improved to a greater extent with the devised information fusion strategy.

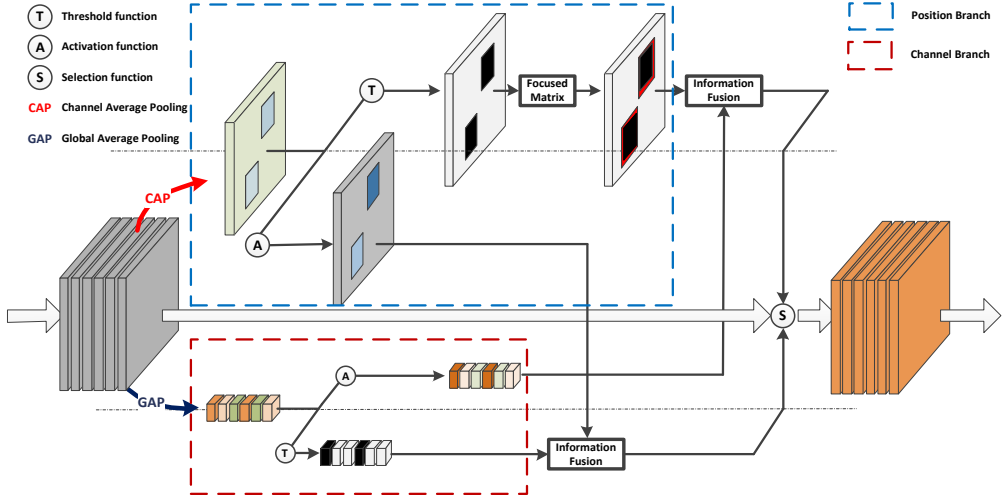

Fig 2: The overall structure of proposed DFM. It contains position and channel branches, focused matrix, and information fusion blocks. The computation process for each map is elaborated in the subsections.

## Weakly Supervised Object Localization

WSOL aims to solve expensive annotation problem by using image-level labels. The multiple-instance learning with CNN features is employed to localize objects in some researches (Cinbis et al. 2015; Gao et al. 2018). A method for self-taught object localization involving masking out image regions to identify the regions is proposed (Bazzani et al. 2016). CAM is employed to identify the discriminative image regions in a single forward (Zhou et al. 2016). A method to tweak the images by randomly hiding patches so that more object parts could be recognized (Singh and Lee 2017). Wei et al., Zhang et al., and Li et al. both replace the most discriminative regions as masks and look for the rest parts of object in the following network (Wei et al. 2017; Zhang et al. 2018; Li et al. 2018). A dropout layer is designed to randomly select importance map and mask map, in order to achieve trade-off between CNN classification and localization accuracy (Choe and Shim 2019). Our work DFM takes the channel information into consideration, not limited on the position aspect. The most effective part is the information fusion between *channel branch* and *position branch*, enhancing classification and localization accuracy simultaneously.

## Dual-Attention Focused Module

The overall structure of DFM is shown in Fig. 2. It is concise and insertable without extra trainable layers. There are two parallel attention branches with implicit information communication. The neighbor focused matrix is embedded in the position mask branch.

## Channel Branch and Position Branch

The *Channel Branch* and *Position Branch* are computational units which introduce channel and spatial information. As they are fully self-attention and have no kernel restriction, these two branches could be employed on any intermediate layer of CNN model. Assume a feature map $\mathbf{F_{in}} \in \mathbb{R}^{c \times h \times w}$ as the input, the channel attention map $\mathbf{C_A} \in \mathbb{R}^{c \times 1 \times 1}$ and position attention map $\mathbf{P_A} \in \mathbb{R}^{1 \times h \times w}$ are respectively computed by the GAP and Channel Average Pooling (CAP).

$$\mathbf{C_A} = \text{GAP}(\mathbf{F_{in}}) = [l_1, l_2, ..., l_c] \\ = [\sum_{i=1}^{w}\sum_{j=1}^{h} F_{1,i,j}, \sum_{i=1}^{w}\sum_{j=1}^{h} F_{2,i,j}, ..., \sum_{i=1}^{w}\sum_{j=1}^{h} F_{c,i,j}]/(w \times h) \quad (1)$$

$$\mathbf{P_A} = \text{CAP}(\mathbf{F_{in}}) = \begin{bmatrix} k_{11} & \cdots & k_{1w} \\ \vdots & \cdots & \vdots \\ k_{h1} & \cdots & k_{hw} \end{bmatrix} = [\sum_{i=1}^{c} F_i]/c \quad (2)$$

where the GAP() represents GAP function and CAP() indicates CAP function. As the pixel value in each point corresponds to the contribution score to network classification, we could recognize the most discriminative channels and positions whose average values rank top list. Then the attention map $\mathbf{C_A} \in \mathbb{R}^{c \times 1 \times 1}$ and $\mathbf{P_A} \in \mathbb{R}^{1 \times h \times w}$ are delivered to following steps to calculate the four key components, enhancement maps $\mathbf{C_E}, \mathbf{P_E}$ and the mask maps $\mathbf{C_M}, \mathbf{P'_M}$.

$$\mathbf{C_E} = \tanh(\mathbf{C_A}) \quad (3)$$

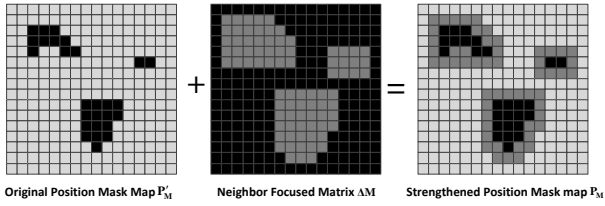

Fig 3: The sample graph using neighbor focused matrix $\Delta \mathbf{M}$ to strengthen surrounding pixel values of the PMM $\mathbf{P}'_\mathbf{M}$. The black color represents zero. Better view with zoom in.

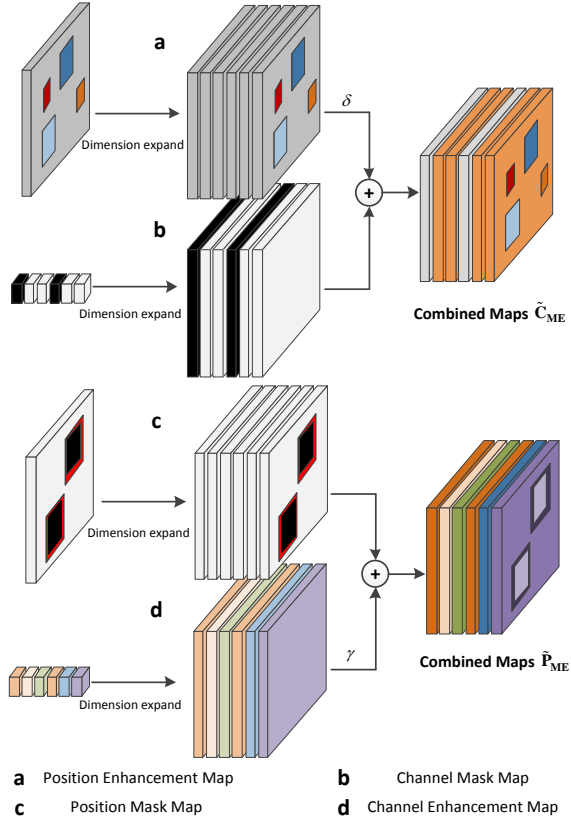

| | |
|---|---|
| **a** Position Enhancement Map | **b** Channel Mask Map |
| **c** Position Mask Map | **d** Channel Enhancement Map |

Fig 4: The detailed structure and example diagram of information fusion block $\mathbf{P}'_\mathbf{M}$. The one above is the fusion process between PEM and CMM. The PEM brings in the abandoned information of CMM and supplies the position information to channel branch. The one below shares a similar principle for information fusion.

$$\mathbf{P_E} = \tanh(\mathbf{P_A}) \quad (4)$$

$$\mathbf{C_M} = \text{Threshold}(\mathbf{C_A}) = \begin{cases} l_i = 0, \text{if } l_i \geq \alpha \cdot l_{max} \\ l_i = 1, else \end{cases} \quad (5)$$

$$\mathbf{P}'_\mathbf{M} = \text{Threshold}(\mathbf{P_A}) = \begin{cases} k_{i,j} = 0, \text{if } k_{i,j} \geq \beta \cdot k_{max} \\ k_{i,j} = 1, else \end{cases} \quad (6)$$

where $\alpha, \beta$ are hyperparameters which control the threshold value. The larger $\alpha, \beta$ imply the more channels and positions will be included as the mask map. We employ tanh activation function here as the large slopes provide more significant differences, and then the discriminative regions are more highlighted compared to other places.

### Neighbor Focused Matrix

The neighbor focused matrix is an auxiliary tool to rationalize the PMM $\mathbf{P}'_\mathbf{M}$, which utilize the principle that pixels of one object are neighbors. As seen in Fig. 3, the matrix strengthens the weights surrounding the mask regions. Suppose the mask regions are $(i, j) \in Mask$, The computation process of the matrix $\Delta \mathbf{M}$ is as follows.

$$\Delta \mathbf{M} = \omega \cdot (P_{s,k} | s \in [i-1, i+1]; k \in [j-1, j+1]) \quad (7)$$

where $\omega$ notes the weights of the neighbor focused matrix. The value of $\omega$ cannot be too large in case the subsidiary attention in the following steps tightly locate the most discriminative regions. For instance, if the most discriminative region for an alligator image is its head, the subsidiary attention hardly reaches the tail when $\omega$ is set too large. Of course, the $\omega$ should not be too tiny to lose its function. For each mask position in $\mathbf{P}'_\mathbf{M}$, the surrounding pixels values are then strengthened as follows.

$$\mathbf{P_M} = \mathbf{P}'_\mathbf{M} + \Delta \mathbf{M} \quad (8)$$

$\mathbf{P_M}$ is the strengthened mask map which highlights the surrounding pixels of mask regions to some degree. This helps avoid the following attention scatter and transfer, thereby improve the classification and localization performance.

### Information Fusion

We complete the information fusion with the enhancement maps $\mathbf{C_E}, \mathbf{P_E}$ and the mask maps $\mathbf{C_M}, \mathbf{P_M}$, as shown in Fig. 4. As we mentioned in the Introduction, the mask maps $\mathbf{C_E}, \mathbf{P_E}$ directly delete all the selected regions information, thereby losing all the implicit reference for the rest parts detection of the object. This issue is extremely aggravated when the rest parts of object are not informative as we suppose. In this case, the subsidiary attention map will be scattershot and easily transfer to the background, which absolutely declines classification and localization accuracy.

To solve this problem, we consider supplementing the implicit information to mask maps to some degree. And we found the enhancement maps $\mathbf{C_E}, \mathbf{P_E}$ are the preferred candidates for this task. Firstly, $\mathbf{C_E}, \mathbf{P_E}$ have the abandoned regions or channels information of $\mathbf{C_M}, \mathbf{P_M}$, so they are capable to alleviate attention map scattershot problem. Second, the $\mathbf{C_E}, \mathbf{P_E}$ themselves contain the important information in channel and position, which enhance the classification and localization accuracy additionally. We firstly align the dimensions of the four key maps.

Table 1: The ablation experiments of DFM components, dual branch, fusion strategy, and focused matrix. The baseline is ResNet50 and the contents in subscript brackets represent the differences compared with CAM results. The positive value indicates improvement and negative implies decay. The bold mark shows the largest value in this column. As we follow the final *CAM* generation as (Zhou et al. 2016), we tick the blank all the time except for the baseline.

| Components | | | | | ILSVRC 2016 | | CUB-200-2011 | |
|---|---|---|---|---|---|---|---|---|
| CAM | Channel | Position | Fusion Strategy | Focused matrix | *Top-1 Clas* | *Top-1 Loc* | *Top-1 Clas* | *Top-1 Loc* |
| | | | | | 76.87 | - | 79.76 | - |
| √ | | | | | 76.06 | 45.35 | 78.49 | 41.17 |
| √ | √ | | | | $76.94_{(+0.88)}$ | $45.92_{(+0.57)}$ | $80.53_{(+2.04)}$ | $46.73_{(+5.56)}$ |
| √ | | √ | | | $75.61_{(-0.45)}$ | $47.37_{(+2.02)}$ | $77.97_{(-0.52)}$ | $50.04_{(+8.87)}$ |
| √ | √ | √ | | | $76.22_{(+0.16)}$ | $46.09_{(+0.74)}$ | $78.67_{(+0.18)}$ | $48.50_{(+7.33)}$ |
| √ | √ | √ | √ | | $77.13_{(+1.07)}$ | $48.82_{(+3.47)}$ | $\mathbf{81.22}_{(+2.73)}$ | $55.42_{(+14.25)}$ |
| √ | √ | √ | √ | √ | $\mathbf{77.76}_{(+1.70)}$ | $\mathbf{49.61}_{(+4.26)}$ | $81.06_{(+2.57)}$ | $\mathbf{56.14}_{(+14.97)}$ |

$$\tilde{\mathbf{C}}_{\mathbf{E}} = [\mathbf{C}_{\mathbf{E},i,j} = \mathbf{C}_{\mathbf{E}} | \forall i \in [1,h]; j \in [1,w]] \quad (9)$$

$$\tilde{\mathbf{C}}_{\mathbf{M}} = [\mathbf{C}_{\mathbf{M},i,j} = \mathbf{C}_{\mathbf{M}} | \forall i \in [1,h]; j \in [1,w]] \quad (10)$$

$$\tilde{\mathbf{P}}_{\mathbf{E}} = [\mathbf{P}_{k,\mathbf{E}} = \mathbf{P}_{\mathbf{E}} | \forall k \in [1,c]] \quad (11)$$

$$\tilde{\mathbf{P}}_{\mathbf{M}} = [\mathbf{P}_{k,\mathbf{M}} = \mathbf{P}_{\mathbf{M}} | \forall k \in [1,c]] \quad (12)$$

After dimension expanding, the four maps have coincident dimensions $\mathbf{C}, \mathbf{P} \in \mathbb{R}^{c \times h \times w}$. We conduct the information communication between *channel branch* and *position branch* as follows.

$$\tilde{\mathbf{C}}_{\mathbf{ME}} = \delta \cdot \tilde{\mathbf{P}}_{\mathbf{E}} + \tilde{\mathbf{C}}_{\mathbf{M}} \quad (13)$$

$$\tilde{\mathbf{P}}_{\mathbf{ME}} = \gamma \cdot \tilde{\mathbf{C}}_{\mathbf{E}} + \tilde{\mathbf{P}}_{\mathbf{M}} \quad (14)$$

where $\tilde{\mathbf{C}}_{\mathbf{ME}}$ represents a combination between the CMM $\tilde{\mathbf{C}}_{\mathbf{M}}$ and the PEM $\tilde{\mathbf{P}}_{\mathbf{E}}$. $\tilde{\mathbf{P}}_{\mathbf{ME}}$ defines a fusion between the CMM $\tilde{\mathbf{P}}_{\mathbf{M}}$ and the CEM $\tilde{\mathbf{C}}_{\mathbf{E}}$. The two hyperparameters $\delta$ and $\gamma$ are utilized to adjust a reasonable ratio between the two fusion components. Considering the advantages of employing $\tilde{\mathbf{P}}_{\mathbf{E}}$ instead of $\tilde{\mathbf{C}}_{\mathbf{E}}$ in formula (13), the intuitive difference in the process is shown in Fig. 4. The $\tilde{\mathbf{P}}_{\mathbf{E}}$ brings in the position importance information while $\tilde{\mathbf{C}}_{\mathbf{E}}$ just decay the mask map $\tilde{\mathbf{C}}_{\mathbf{M}}$ function in proportion. It is the same in formula (14).

After the information fusion, the two combined map $\tilde{\mathbf{C}}_{\mathbf{ME}}$ and $\tilde{\mathbf{P}}_{\mathbf{ME}}$ are randomly fed back to the input feature maps. This is the final step of DFM to complete the WSOL task. The random mechanism avoids the simultaneous appearance of the enhancement map and the mask map which are from the same branch. If it happens, the functions of enhancement map will be eliminated by the mask map, as there are troughs in mask regions to flatten the highlight regions in enhancement map. And the information from channel and position could not be fused to improve WSOL performance.

$$\mathbf{F}_{\mathbf{Module}} = \text{random}(\tau, \tilde{\mathbf{C}}_{\mathbf{ME}}, \tilde{\mathbf{P}}_{\mathbf{ME}}) \quad (15)$$

$$\mathbf{F}_{\mathbf{out}} = \mathbf{F}_{\mathbf{in}} + \mathbf{F}_{\mathbf{Module}} \quad (16)$$

where the $\mathbf{F}_{\mathbf{Module}}$ represents the output of our proposed DFM, $\tau$ is the selection probability of $\tilde{\mathbf{P}}_{\mathbf{ME}}$, and the $\mathbf{F}_{\mathbf{out}}$ is the revised feature maps which is delivered to the next layer computation. With the two branches, neighbor focused matrix, and information fusion strategy, the missing information of mask maps is compensated in a simple but effective way. The enhancement map from one branch offers the other branch abundant information, including the implicit hint of the most discriminative part and importance information for better classification and localization performance.

## Experiment

### Experiment setups

**Datasets:** We complete experiments on the two public datasets ILSVRC 2016 (Russakovsky et al. 2015) and CUB-200-2011 (Wah et al. 2011) to evaluate the DFM performance in WSOL task. There are about 1.2 million images in ILSVRC 2016 for training and 50,000 images for testing, which contains 1,000 categories. CUB-200-2011 has 5,994 images for training and 5,794 for testing, which consists of 200 kinds of birds. The challenging for ILSVRC WSOL is the variety and the complex background, and the difficulty in CUB-200-2011 relies on the elusive difference and various attitudes of the birds.

**Metrics:** According to the recommendation in (Russakovsky et al. 2015) and state-of-the-art methods (Wei et al. 2018; Zhang et al. 2018; Zhang et al. 2018; Choe and Shim 2019), we utilized the Top-1 classification accuracy (*Top-1 Clas*) and Top-1 localization accuracy (*Top-1 Loc*) as the evaluation metrics. *Top-1 Clas* represents the ratio of correct classification prediction. *Top-1 Loc* is ratio of accurate classification and correct object localization when the intersection over union (IoU) is not less than 50%.

Table 3: Comparison between DFM with state-of-the-art WSOL method. The experiment utilizes the VGG16, ResNet50, ResNet101, and MobileNetV1 as the baseline. The bold mark means the largest value in this column. We also calculate the difference value between our DFM and the highest value in other state-of-the-art methods. The value is listed in the subscript brackets.

| Backbone | Method | ILSVRC 2016 | | CUB-200-2011 | |
| --- | --- | --- | --- | --- | --- |
| | | *Top-1 Clas* | *Top-1 Loc* | *Top-1 Clas* | *Top-1 Loc* |
| ResNet50 | CAM (2016) | 76.06 | 45.35 | 78.49 | 41.17 |
| | DFM | **77.76**$_{(+1.70)}$ | **49.61**$_{(+4.26)}$ | **81.06**$_{(+2.57)}$ | **56.14**$_{(+14.97)}$ |
| ResNet101 | CAM (2016) | 76.93 | 46.16 | 78.26 | 40.02 |
| | DFM | **77.52**$_{(+0.59)}$ | **50.65**$_{(+4.49)}$ | **81.52**$_{(+3.26)}$ | **54.68**$_{(+14.66)}$ |
| VGG16 | CAM (2016) | 66.65 | 42.76 | 67.79 | 34.56 |
| | ACoL (2018) | 67.50 | 45.83 | 71.90 | 45.92 |
| | ADL (2019) | **69.48** | 44.92 | 65.27 | 52.36 |
| | DFM | 68.60$_{(-0.88)}$ | **47.41**$_{(+1.58)}$ | **72.52**$_{(+0.62)}$ | **55.84**$_{(+3.48)}$ |
| MobileNetV1 | CAM (2016) | 68.38 | 41.66 | 71.94 | 43.70 |
| | HaS (2017) | 67.48 | 41.87 | 66.64 | 44.67 |
| | ADL (2019) | 67.77 | 43.01 | 70.43 | 47.74 |
| | DFM | **68.63**$_{(+0.25)}$ | **44.19**$_{(+1.18)}$ | **72.37**$_{(+0.43)}$ | **49.13**$_{(+1.39)}$ |

**Implementation:** We evaluated the DFM capability with the baselines VGG16 (Simonyan and Zisserman 2015), Resnet50, Resnet101 (He et al. 2016), and MobileNetV1 (Howard et al. 2017). We also plug our module in the bottleneck between stages (Choe and Shim 2019. The CAM and bounding box are obtained as the same as the previous work (Zhou et al. 2016). We employ the ILSVRC pretrained models and fine-tune them after adding in our module. The hyperparameters in our work is tested for good results. During the two branches computation of ResNet50, the mask thresholds are set as $\alpha = 0.85$, $\beta = 0.95$, and strengthen ratio $\omega = 0.15$. In the information fusion, $\delta = 0.6$ and $\gamma = 0.4$. For maps selection, $\tau = 0.70$, i.e., there is 70% probability to choose $\tilde{P}_{ME}$. The DFMs in ResNet50 are inserted following the conv5_3 and conv4_1. We adopt the PyTorch as framework and train our model on NVIDIA GeForce TITAN V GPU.

### Ablation Study

We detail to evaluate the importance of each branch, information fusion, and neighbor focused matrix. Additionally, we test the model performance under various hyperparameters setting. Here we take the ResNet50 as the network baseline.

First, the importance of each component is listed in Table 1. The classification result of CAM is slightly lower than baseline as the final layers have been replaced, and the localization accuracy is limited as only the most discriminative regions are detected. The position branch contributes a lot to the localization accuracy while the channel branch increases the classification accuracy. We set a normal fusion group ($\tilde{C}_M$ with $\tilde{C}_E$, $\tilde{P}_M$ with $\tilde{P}_E$) to compare with the proposed strategy. The dual-attention module achieves superior performance in *Top-1 Loc* and *Top-1 Clas*. The experiment effect of neighbor focused matrix varies in a different situation. Basically, it leads to floating improvement in localization.

Table 2: Hyperparameters adjustment in information fusion. Backbone is ResNet50 and test on CUB-200-2011. Results are *Top-1 Loc*. The other hyperparameters are set as Implementation.

| | | $\gamma$ | | | |
| --- | --- | --- | --- | --- | --- |
| | | 0.9 | 0.5 | 0.4 | 0.1 |
| $\delta$ | 0.9 | 48.77 | 52.13 | 53.06 | 49.11 |
| | 0.7 | 52.61 | 55.09 | 55.75 | 50.40 |
| | 0.6 | 51.90 | 55.26 | 56.14 | 50.32 |
| | 0.1 | 49.45 | 51.94 | 52.38 | 48.76 |

After analyzing the significance of DFM components, we evaluate the fusion parameters in WSOL task. According to the results in Table 2, the value of the compensation information from the enhancement maps should be discounted. Otherwise, the estimated object region would be small as it loses the idea of the most discriminative regions erasing.

### Comparing with State-of-the-art Work

We compare the proposed DFM with the state-of-the-art WSOL methods including CAM (Zhou et al. 2016), ACoL (Zhang et al. 2018), HaS (Singh and Lee, 2017), ADL (Choe and Shim 2019). In addition, we also take strongly supervised methods into consideration, such as CBAM (Woo et al. 2018) and ABN (Fukui et al. 2019).

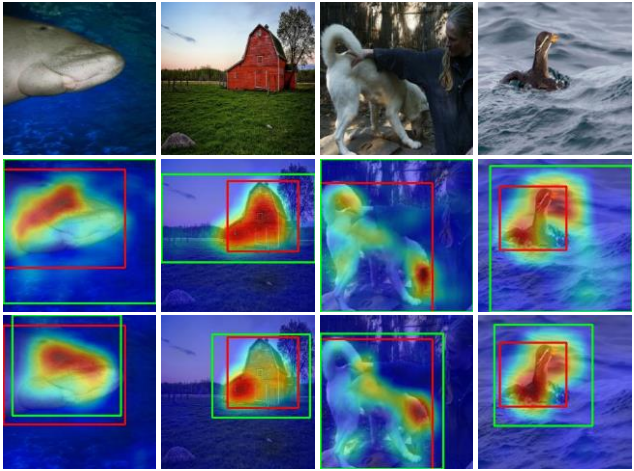

Fig 5: Some exemplar visualization results of ADL and DFM methods. Red boxes represent ground truths and green ones are estimated results. Our DFM results are in the third row.

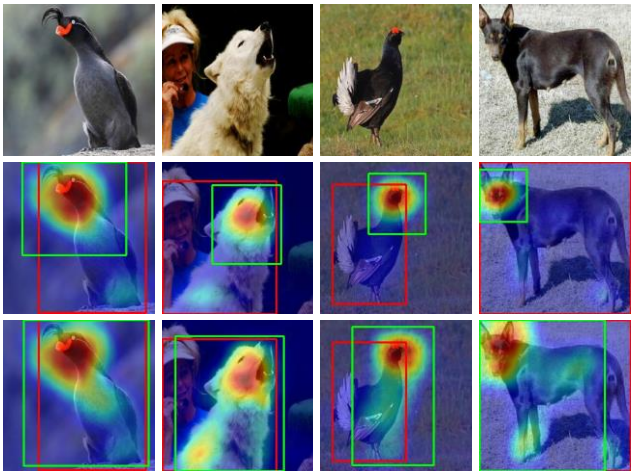

Fig 6: The comparison between CAM and DFM methods.

**WSOL:** The quantitative comparison result in WSOL is shown in Table 3. DFM achieves outperforming performance in both *Top-1 Loc* and *Top-1 Clas* with different backbones. In the localization task, DFM improves the accuracy by 1.58%, 3.48%, 1.18%, and 1.39% compared with the highest *Top-1 Loc* values using VGG16 and MobleNetV1 respectively, which also verify the generality of DFM. For ResNet50 and ResNet101, the hyperparameters are shared which also obtain more than 4% and 14% *Top-1 Loc* enhancement in ILSVRC2016 and CUB-200-2011 respectively. We also acquire a satisfactory *Top-1 Clas* in the experiment, especially with the ResNet. The visualization results are shown in Fig. 5 and Fig. 6. We intuitively learn the DFM promote the entire object detection and avoid the attention regions being rambling.

**Strongly Supervised Classification:** We also compare our results with some strong supervised methods, CBAM (Woo et al. 2018) and ABN (Fukui et al. 2019). The results are listed in Table 4.

Table 4: *Top-1 Clas* comparison between our DFM with state-of-the-art strong supervised methods which utilize the attention information. Bold means the highest value and the difference value is contained in the subscript brackets. In the *task*, *ssol* represents strongly supervised object localization, *wsol* indicates weakly supervised object localization, and *ssc* means strongly supervised classification.

| Backbone | Method | task | ILSVRC 2016 *Top-1 Clas* |
|---|---|---|---|
| ResNet50 | CBAM (2018) | ssol | 77.34 |
| | DFM | wsol | **77.76**$_{(+0.33)}$ |
| ResNet101 | CBAM (2018) | ssol | **78.49** |
| | DFM | wsol | 77.52$_{(-0.97)}$ |
| VGG16 | ABN (2019) | ssc | **68.80** |
| | DFM | wsol | 68.60$_{(-0.20)}$ |
| MobleNetV1 | CBAM (2018) | ssol | **70.99** |
| | DFM | wsol | 68.63$_{(-2.36)}$ |

CBAM is a strongly supervised object localization method, so it has position annotation to provide abundant information for localization and classification, so it behaves better with MobleNetV1 and ResNet101. However, DFM result using ResNet50 is a little higher than CBAM. ABN is designed for strongly supervised classification, which requires abundant overheads parameters. The accuracy between ABN and DFM are approximately the same. Based on these results, the DFM has shown considerable capability and potential in object classification compared with the strongly supervised learning algorithms.

Verified by the experiment results, we are capable to know the classification and localization accuracies of DFM outperform the state-of-the-art WSOL methods, which also closes to the strongly supervised level. Considering that DFM structure is concise but effective, its variants with the structural reinforcement are supposed to be researched for further WSOL performance enhancement.

## Conclusion

In this work, we have proposed a dual-attention focused module to enhance the classification and localization accuracy in WSOL task. We analyzed the two drawbacks of the most discriminative regions erasing strategy at the present. To address this problem, the channel branch and position branch are introduced to extract self-attention maps which cover the comprehensive information in channel and position. We devise a maps fusion policy, integrating the enhancement map with mask map, to compensate the erased information, thereby offering implicit hint and importance information from the other branch for following attention regions detection and object classification. Additionally, the neighbor focused matrix is proposed to highlight the surrounding pixels of erased regions to prevent attention from transferring to the non-object region. The proposed DFM needs no extra overheads parameters and easily plugs into various backbones. The experiments verify DFM enhances the performance in WSOL, both in the classification and localization, which outperforms the state-of-the-art methods.